\documentclass[runningheads]{llncs}
\usepackage{graphicx}
\usepackage{amsmath,amssymb}

\begin{document}

\title{On the utility and protection of optimization with differential privacy and classic regularization techniques}
\titlerunning{Utility and protection of differential privacy and regularization techniques}
%
\author{Eugenio Lomurno \and Matteo Matteucci}
\authorrunning{E. Lomurno et al.}
%
\institute{Politecnico di Milano, Milano, Italy\\
\email{eugenio.lomurno@polimi.it, matteo.matteucci@polimi.it}}
\maketitle

\begin{abstract}
Nowadays, owners and developers of deep learning models must consider stringent privacy-preservation rules of their training data, usually crowd-sourced and retaining sensitive information.
The most widely adopted method to enforce privacy guarantees of a deep learning model nowadays relies on optimization techniques enforcing differential privacy.
According to the literature, this approach has proven to be a successful defence against several models' privacy attacks, but its downside is a substantial degradation of the models' performance.
In this work, we compare the effectiveness of the differentially-private stochastic gradient descent (DP-SGD) algorithm against standard optimization practices with regularization techniques. 
We analyze the resulting models' utility, training performance, and the effectiveness of membership inference and model inversion attacks against the learned models.
Finally, we discuss differential privacy's flaws and limits and empirically demonstrate the often superior privacy-preserving properties of dropout and l2-regularization.
\end{abstract}

\section{Introduction}

In recent times, the number and variety of machine learning applications have noticeably grown thanks to the computational progress and to the availability of large amounts of data.
In fact, given the proportionality between the performance of machine learning models and the amount of data fed into them, more and more data has been and is going to be collected to get reliable and accurate results.
This data is usually crowd-sourced and may contain private information.
Thus, it is necessary to guarantee high privacy levels for data owners, concerning both the sharing of their information and the machine learning models trained over it.

In fact, malicious agents can target deep learning models to exploit the information that remains within them after training is complete, even in black-box scenarios.
Among the most famous and dangerous attacks, membership inference~\cite{shokri2017membership} and model inversion~\cite{fredrikson2015model} techniques represent the main threat for machine learning models.
The former aims at guessing the presence of an instance inside the training data of the attacked model, while the second has the goal of reconstructing the input data from the accessible or leaked information.

To reduce the effectiveness of these attacks against artificial neural networks, a widely spread countermeasure taken by model owners is applying differential privacy~\cite{dwork2008differential} in the models' training procedure. 
The main advantage of this solution is the guarantee that the privacy leakage at the end of models' training is limited and measurable. The main downside of this approach
relies in the noise injected in the models during their training to achieve the desired privacy budget level. In fact, noise addition significantly impacts over models' training performance in terms of time and utility.

In this work, we investigate the topic of privacy preservation in deep learning.
We test the effectiveness of the most known implementation of differential privacy training for deep learning models, the differentially private stochastic gradient descent (DP-SGD)~\cite{abadi2016deep}, as a defense mechanism.
We evaluate the validity of this method in terms of protection against model inversion and membership inference attacks. We also measure the impact that DP-SGD has on the model under attack, analyzing both the level of accuracy achieved and the training time required in white-box and black-box scenarios.

We perform the same analysis considering two regularization techniques, i.e., dropout and the l2-regularizer. Their effect on improving model's generalization capability is widely known, and previous works have confirmed an existing connection between privacy attacks' effectiveness and overfitting in the target model~\cite{salem2018ml}. Our work empirically demonstrates how l2 regularization and dropout achieve similar or even better privacy-preserving performance with respect to DP-SGD training while preserving models utility and training time efficiency.
To the best of our knowledge, this is the first work to systematically analyze the impact of different levels of differential privacy and regularization techniques in terms of models' accuracy, training time, and resistance to membership inference and model inversion attacks.

\section{Privacy threats}
Privacy in machine learning is a much-debated topic in the literature.
If, on the one hand, deep learning algorithms efficiently learn from data how to solve complex tasks, on the other hand, it has been demonstrated how these models exhibit vulnerabilities to malicious attacks despite their complexity~\cite{dwork2015robust,song2019membership,hu2021membership}.
The targets of these attacks are various and characterized by different risk levels from a user perspective.

We differentiate the scenarios in which an attack occurs depending on the extent of the adversarial knowledge, that is, the ensemble of information concerning the model and the data under attack at the disposal of the attacker. From this point of view, we distinguish the threat scenarios into two types: black-box and white-box. In a black-box scenario, the adversary knows only the target model elements available to the public, such as prediction vectors, but has no access to the model structure or information about the training dataset outside its format. In a white-box scenario, the adversary has complete knowledge of the target model and knows the data distribution of the training samples.
Among the most famous and dangerous attacks, we focus our attention on membership inference and model inversion attack families~\cite{al2019privacy}.

\subsection{Membership inference attacks}
Membership inference attacks take as input a sample and try to determine if it belongs or not to the training dataset of the model under attack. The most common design paradigm involves the use of shadow models and meta-models, also known as attack models~\cite{shokri2017membership}.
The basic idea behind this white-box approach is to train several shadow models that imitate the behavior of the target on surrogate or shadow datasets. Shadow datasets must contain samples with the same format and similar distribution to the training data of the target model. After the training of shadow models is complete, their outputs and the known labels from the shadow datasets form the attack dataset. 
This dataset is used to train a meta-model which learns to make membership inference based on the shadow models' results.

The main limitation of this approach is represented by the mimic capabilities of the shadow models with respect to the target model and the strong assumptions related to the adversarial knowledge of both the target model structure and the training data distribution.
To overcome these issues, Salem \textit{et al.}~\cite{salem2018ml} proposed three different attacks considering scenarios with more relaxed assumptions on the adversarial knowledge. The first two approaches maintain the idea of shadow models, while the third proposal abandons the shadow model paradigm in favor of a threshold-based attack. In this approach, the attacking model is a simple binary classifier that takes the highest posterior from the prediction vector of the target model and compares it against a given threshold. If its value is greater than the threshold, the input sample obtained from that output is considered a member of the training dataset. The advantages of this approach concern the complete independence from the target model and its training data and the elimination of the overhead costs due to the design of shadow models and the creation of suitable shadow datasets. 
Besides, it requires no training of the attack model.
A novel type of membership inference attack, named BlindMI, has recently been proposed by Hui \textit{et al.}~\cite{hui2021practical} to probe the target model and extract membership semantics via differential comparison and data generation.

\subsection{Model inversion attacks}
Model inversion attacks try to reconstruct training samples of the attacked model starting from environmental elements known by the attacker. The first designs of reconstruction attacks assumed a white-box scenario in which the adversary knows the output label, the prior distribution of features for a given sample, and has complete access to the model. With these assumptions, the attacker estimates the sensitive attributes' values that maximize the probability of observing the known model parameters. These first forms of attacks are referred to as Maximum a-posteriori (MAP) attacks~\cite{fredrikson2014privacy}. However, they were soon abandoned because their performance degrades as the feature space to reconstruct grows.
To overcome this limitation, Frederikson \textit{et al.}~\cite{fredrikson2015model} proposed to formulate the attack as an optimization problem where the objective function depends on the target model output. Starting from assumptions similar to those considered in MAP attacks, the attacker reconstructs the input sample through gradient descent in the input space. 
This attack can be performed in white and black box scenarios, depending on whether the attacker has access to the intermediate maps of the target model or its prediction vector respectively.

Yang \textit{et al.}~\cite{yang2019neural} proposed a black-box attack where the adversarial does not know any detail about the model and its training data. Instead, it knows the generic training data distribution and the output format of the model, i.e., the prediction vector.
Zhang \textit{et al.}~\cite{zhang2020secret} designed a novel solution that involves the use of a generative adversarial network (GAN)~\cite{goodfellow2014generative} to learn the training data representation, exploiting the properties of this kind of network to increase the feasibility of the attack.
Zhao \textit{et al.}~\cite{zhao2021exploiting} exploited information provided by artificial intelligence explainability tools to achieve high fidelity reconstruction of the target model input data. The authors focus on understanding which explanations are more useful for the attacker, measuring which of them leaks higher information amounts about target data.

Lim and Chan~\cite{lim2021gradient} exploited a privacy-breaking algorithm in a federated learning scenario to reconstruct users' input data from their leaked gradients. The idea is to generate dummy gradients from a randomly initialized dummy input and compute the loss between these and the true ones. The loss is used to update the dummy input to reduce the distance with respect to real input data according to Geiping \textit{et al.}~\cite{geiping2020inverting}.
Finally, Yin \textit{et al.}~\cite{yin2021see} designed GradInversion, an algorithm able to recover with great precision single images from deep networks gradients trained with large batch sizes.
The first step of this approach is to convert the input reconstruction problem into an optimization task, where, starting from random noise, synthesized images are exploited to minimize the distance between the gradients of these and the real gradients provided by the environment. Then, the core of the algorithm lies in a batch-wise label restoration method, together with the use of auxiliary losses that aim to ensure fidelity and group consistency regularization to the final result.

\section{Privacy defences}
With the spread of machine learning as a service, the attack surface for the world of machine learning has undergone rapid growth. Nowadays, machine learning threats and defences are involving disparate scenarios and techniques~\cite{mothukuri2021survey}.
Differential privacy~\cite{dwork2008differential} represents the most proposed technique to guarantee protection and ensure data owners' privacy.

Differential privacy is devised as an effective privacy guarantee for algorithms that work with aggregated data.
It was initially proposed in the domain of database queries, defining the concept of adjacent databases as two sets that differ in a single entry.
More formally, a randomized mechanism \emph{M: $D \rightarrow R$} with domain \emph{D} and range \emph{R} satisfies $\varepsilon$-differential privacy if for any two adjacent inputs $\emph{d, d'} \in \emph{D}$ and for any subset of outputs $\emph{S} \subseteq \emph{R}$ it holds that:

\begin{equation} 
    Pr[\emph{M(d)} \in \emph{S }] \leq \emph{e}^\varepsilon Pr[\emph{M(d')} \in \emph{S }]. 
\end{equation}

The parameter $\varepsilon$ is called privacy budget because it represents how much information leakage can be afforded in a system. The lower the $\varepsilon$ value, the stricter the privacy guarantee.
Differential privacy represents a significant development in the field of privacy-preserving techniques because it guarantees three properties that are very useful, namely: composability, group privacy, and robustness to auxiliary information. 
\begin{itemize}
    \item Composability concerns the possibility of having a composite mechanism so that, if each of its components is differentially private, then is also the overall mechanism itself. This property stays true for sequential and parallel compositions. 
    \item Group privacy assures that if the dataset contains correlated data, like the ones provided by the same individual, the privacy guarantee degrades gracefully and not abruptly. 
    \item Robustness to auxiliary information guarantees that the privacy level assured by the theory stands regardless of the knowledge available to the adversary.
\end{itemize}

The main theoretical issue related to this definition of differential privacy relies in its rigor.
In fact, in order to make it exploitable for real uses it is necessary to relax its constraints.
There exists many formulations that generalize the privacy-budget $\varepsilon$ and provide its relaxation, for instance the $\textit{f}$-differential privacy~\cite{dong2019gaussian} or the concentrated differential privacy~\cite{dwork2016concentrated}.
Among them, the most applied formulations are the ($\varepsilon,\delta$)-differential privacy~\cite{abadi2016deep} and the Rényi differential privacy~\cite{mironov2017renyi}.

($\varepsilon,\delta$)-differential privacy is defined via a randomized mechanism \emph{M: $D \rightarrow R$} with domain \emph{D} and range \emph{R} that satisfies its constraints if for any two adjacent inputs $\emph{d, d'} \in \emph{D}$ and for any subset of outputs $\emph{S} \subseteq \emph{R}$ it holds that:

\begin{equation} 
    Pr[\emph{ M(d)} \in \emph{S }] \leq \emph{e}^\varepsilon Pr[\emph{ M(d')} \in \emph{S }] + \delta,
\end{equation}
where the additive factor $\delta$ represents the probability that plain $\varepsilon$-DP is broken. 
In the case of several mechanism, the composition property still holds, even if in a more complex version that keeps track of the privacy loss accumulated during the execution of each component.
Starting from this composability property, Abadi \textit{et al.}~\cite{abadi2016deep} designed an new function called moments accountant, that computes the privacy cost needed for each access to the data and uses this information to define the overall privacy loss of the mechanism. 
Then, in the same work, Abadi \textit{et al.} defined the so called differentially-private stochastic gradient descent (DP-SGD) that is actually one of the most adopted optimizers to implement differential privacy.

The Rényi Differential Privacy is instead another form of relaxation of differential privacy based on the concept of Rényi divergence.
Given that for two probability distributions \emph{P} and \emph{Q} defined over \emph{R}, the Rényi divergence of order $\alpha > 1$ defined as:

 \begin{equation} \label{eqn:renyi_div}
    \emph{D}_\alpha(\emph{P}\parallel\emph{Q}) \triangleq \frac{1}{\alpha-1} \log E_{x\sim\emph{Q}} \left(\frac{\emph{P(x)}}{\emph{Q(x)}}\right)^\alpha. 
\end{equation}
The relationship between the differential privacy formulation and the Rényi divergence can be defined by a randomized mechanism M: $D \rightarrow R$ that is $\varepsilon$-differentially private if and only if its distribution over any pair of adjacent inputs $d, d' \in \emph{D}$ satisfies:

\begin{equation} 
    \emph{D}_\infty(\emph{M(d)}\parallel\emph{M(d')}) \leq \varepsilon. 
\end{equation}

Putting all together, it is possible to define the ($\alpha, \varepsilon$)-Rényi differential privacy as a randomized mechanism M: $D \rightarrow R$ is said to have $\varepsilon$-Rényi differential privacy of order $\alpha$ if for any adjacent $d, d' \in \emph{D}$ it holds that:

\begin{equation} \label{eqn:rdp}
    \emph{D}_\alpha(\emph{M(d)}\parallel\emph{M(d')}) \leq \varepsilon. 
\end{equation}
It is demonstrated that the three properties of composability, robustness to auxiliary information, and group privacy are still valid for ($\alpha, \varepsilon$)-Rényi differential privacy.

Thus, differential privacy turns out to be the most adopted privacy-preserving technique for its guarantees and its quantification of the privacy-budget \cite{lomurno2021generative,jordon2018pate,wei2020federated,mothukuri2021survey}.
However, Bagdasaryan \textit{et al.}~\cite{bagdasaryan2019differential} argue its negative impact when applied via DP-SGD in terms of performance loss, especially for low $\varepsilon$ values.
Due to its implementation, more privacy guarantees from DP-SGD mean higher noise injections during the training procedure as well as fewer training iterations for the model.
This is coherent with the results of Salem \textit{et al.}~\cite{salem2018ml} work, in which is demonstrated the proportionality between deep learning models' overfitting and their vulnerability to membership inference attacks. This makes sense, on one hand, concerning that DP-SGD noise addiction can be seen as a regularization form, and on the other hand, recalling that differential privacy can be interpreted also as a direct countermeasure to membership inference attacks.

Besides, other works have tried to mitigate these threats with different techniques. 
For instance Jain \textit{et al.}~\cite{jain2015drop} discussed the dropout~\cite{srivastava2014dropout} differentially-private properties and its protection against membership inference attacks.
Ermis \textit{et al.}~\cite{ermis2017differentially} defined a form of differential privacy starting from the Bayesian definition of Gaussian dropout.
Differently, Nasr \textit{et al.}~\cite{nasr2018machine} defined an adversarial regularization to protect their models by changing the loss function, while Yang \textit{et al.}~\cite{yang2020defending} implemented a procedure named prediction purification to protect against both membership inference and model inversion again with adversarial learning.

\section{Method}
This work aims to provide an exhaustive and detailed comparison between deep learning models with and without privacy-preservation techniques.
We tested the effectiveness of both DP-SGD and regularization techniques in defending the target model, as well as their impacts over the accuracy and the training time of the target model itself.
To adopt the DP-SGD we relied on the TensorFlow Privacy implementation based on Abadi \textit{et al.} work~\cite{abadi2016deep} which corresponds to the ($\varepsilon,\delta$)-differential privacy version.

\begin{figure}[t]
    \centering
    \includegraphics[width=.4\textwidth]{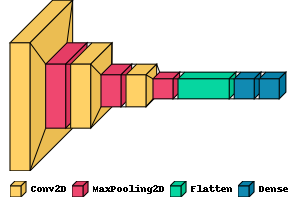}
    \caption{The neural network architecture of the target model.}
    \label{fig:target}
\end{figure}

The target model always has the same architecture despite the training procedure and the regularization techniques adopted, as shown in Figure~\ref{fig:target}. 
For fair comparisons and simplicity, we chose a straightforward convolutional neural network with ReLU activation for hidden layers and a Softmax output activation.

\begin{figure}[t]
    \centering
    \includegraphics[width=1\textwidth]{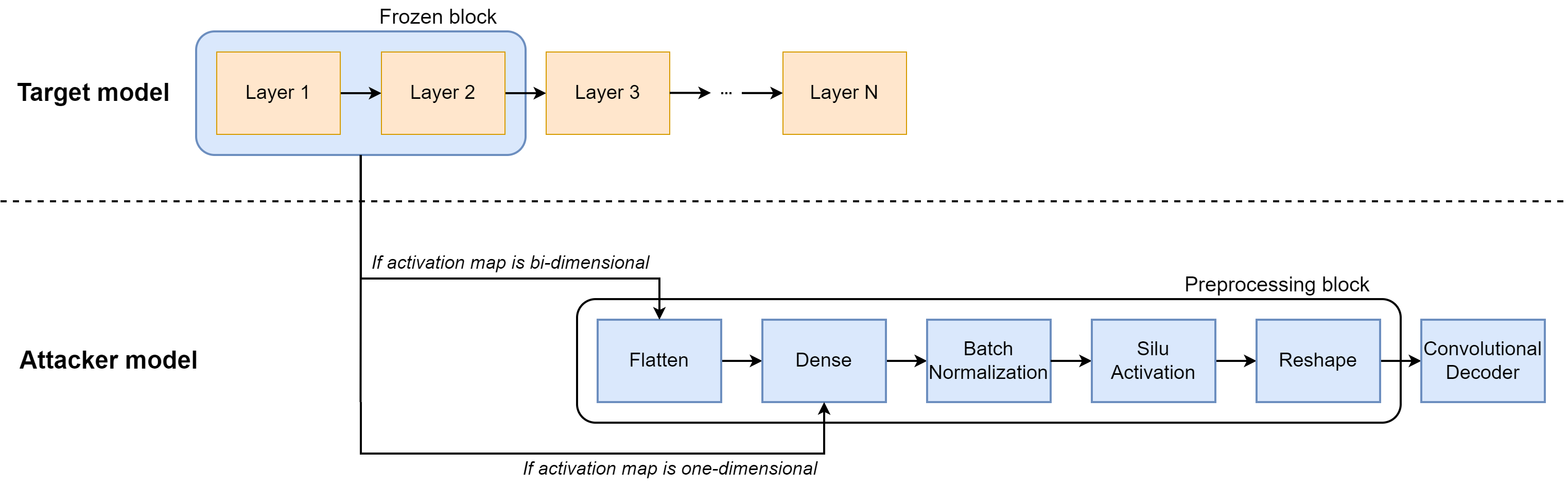}
    \caption{The schema of the proposed model inversion attack. In a white-box scenario the attacker can reconstruct the target model input from an arbitrary activation map (in the example, from the second layer's one). In a black-box scenario, the attacker can only exploit the output of the layer N to invert the target model.}
    \label{fig:model_inversion}
\end{figure}

\begin{figure}[t]
    \centering
    \includegraphics[width=1\textwidth]{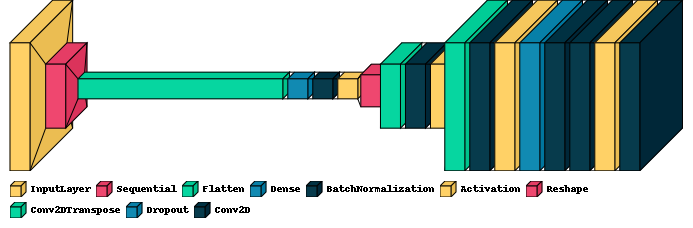}
    \caption{The neural network architecture of the attacker model used for model inversion.}
    \label{fig:adversary}
\end{figure}

To evaluate its resistance to membership inference attacks, we adopted again a TensorFlow Privacy tool implementing two black-box attacks inspired by Salem \textit{et al.} work~\cite{salem2018ml}.
The first is a threshold-based attack, while the second involves the training of a single shadow model and assumes no knowledge about the data distribution.
The attacking tool automatically selects the most effective technique among the aforementioned two.
We measured the Area Under Curve (AUC) as a metric to evaluate the attacks' effectiveness.

Concerning the model inversion attack, we have devised an approach that exploits the activation maps of the target model to reconstruct its training data, as shown in Figure~\ref{fig:model_inversion}.
In detail, after having trained the target model, we select the target layer from which reconstruct the training input, that in a black-box scenario corresponds to the output layer. 
We cut out the subsequent part of the target network and free the weights.
Finally, we use this target network fraction as preprocessing layer for the attacker network to compute the model inversion attack.
A detailed overview of the architecture of the adversary model is shown in Figure~\ref{fig:adversary}, in which the sequential layer represents the frozen fraction of the target model.
After some tuning, we selected mean squared error (MSE) as loss function, SiLU as activation function for all the hidden layers and Adam as optimizer with a learning rate of $10^{-3}$.
The output of the network is a convolutional layer with the same number of channels of the input data of the target and a Sigmoid activation function.
To measure the reconstruction quality of this model inversion attack, we adopted the MSE as evaluation metric.

To recap, in order to introduce differential privacy in the training of the target model is sufficient to substitute its optimizer with the DP-SGD, thus preserving the same structure and number of parameters.
To have a better comprehension of the impact of privacy-preserving techniques, we selected three different privacy budget levels, i.e., $\varepsilon=2$, $\varepsilon=4$, $\varepsilon=8$ keeping the same privacy leak probability among the experiments, i.e., $\delta=10^{-5}$.
We trained a distinct target model with each of these privacy budget.

In parallel, we also trained target models with regularization techniques.
We decided to inspect the impact of the most adopted mechanisms to avoid overfitting, i.e., dropout and weight decay or l2.
In particular, we tested one target model for each of the two techniques and one with both of them together.
The dropout is applied between every weighted layer, i.e., after each convolutional layer and before and after the first dense layer of the target model.
The l2 is applied directly to the last dense layer, i.e., the prediction layer.
\section{Experiments and results}
All experiments described below have been carried out on a system equipped with an Intel(R) Xeon(R) CPU E5-2630 v4 @ 2.20GHz and an Nvidia GeForce GTX TITAN X GPU.
We performed our analysis on three image datasets, i.e., CIFAR-10, MNIST and Fashion-MNIST, each of them pre-processed with a normalization step.

In order to have comparable experiments, we fixed some hyperparameters among the different target model configurations.
The $\varepsilon$ parameter is typically computed a-posteriori, to evaluate the privacy budget achieved by a model with a differentially-private optimizer.
In the formulation provided by Abadi \textit{et al.}~\cite{abadi2016deep}, $\varepsilon$ is function of the number of data samples, the number of micro-batches, the noise multiplier and the number of training epochs.
In our setting, we have fed the differentially-private target models with the same data as the other models, with the same batch size = 200.
In order to maximize the utility, we set the number of micro-batches equal to the number of mini-batches.
Finally, in order to control a-priori the privacy budget, we fixed the number of training epochs to the optimal value found for the target model without privacy-preserving techniques.
In this way, we could control the $\varepsilon$ privacy budget by varying the noise multiplier parameter.
We did similar reasoning concerning the regularization techniques by fixing batch size, the number of data samples, and the optimizer.
After some tuning, we also fixed the dropout rate to 20\% and the l2-weight to 2x$10^{-2}$.
As an optimizer to compare with the DP-SGD we chose the SGD.
For both of them, we tuned the learning rate that converged to the same optimal value of $10^{-1}$.

\begin{table}[t]
    \caption{The accuracy scores of the target model over the test set (higher is better).}
    \centering
    \begin{tabular}{|cccccccc|}
        \hline
        \textbf{  Dataset  } & \textbf{Baseline} & \textbf{  $\varepsilon$ = 2  } & \textbf{  $\varepsilon$ = 4  } & \textbf{  $\varepsilon$ = 8  } & \textbf{  L2  } & \textbf{  Dropout  } & \textbf{  L2+Dropout  }\\
        \hline
        CIFAR-10 & 0.664 & 0.518 & 0.538 & 0.536 & 0.649 & $\mathbf{0.692}$ & 0.648\\
        MNIST & 0.992 & 0.942 & 0.963 & 0.968 & 0.991 & $\mathbf{0.996}$ & 0.994\\
        F-MNIST & $\mathbf{0.900}$ & 0.820 & 0.823 & 0.829 & 0.896 & 0.890 & 0.879\\
        \hline
    \end{tabular}
    \label{tab:dp_test}
\end{table}

Table~\ref{tab:dp_test} shows the results of each target model over the test sets, which confirms the thesis of Bagdasaryan \textit{et al.}~\cite{bagdasaryan2019differential}.
According to their results, we notice a performance drop inversely proportional to the privacy budget $\varepsilon$.
This behavior is reasonable concerning that higher differential privacy levels correspond to higher levels of noise injection during the training.
However, in the most private scenario with $\varepsilon=2$, the target model left from 5\% to 14.6\% of accuracy, while in the other differentially-private cases, the performance loss is slightly lower but still consistent.
Concerning the regularization techniques, the performance is almost identical to the baseline for l2 regularization, while it is ever improved with the dropout application, especially over CIFAR-10. Finally, the combination of l2 and dropout brings to mildly pejorative performance. We argue that this behavior is due to the probable excess of regularization.

Table~\ref{tab:dp_times} shows the time required by each target model configuration to compute the same number of epochs on the proposed datasets. 
In this dimension, all the models trained via DP-SGD completed their training in a largely greater time amount than the one required by the other configurations.
This behavior, which for $\varepsilon=2$ means an increment factor between 41 and 50, is related to the implementation of the DP-SGD.
In fact, to apply the differential privacy definition and compute its budget, it is necessary to expand the dimension of the tensor fed in the network to include the micro-batch channel. 
The number of micro-batches, constrained between one and the number of mini-batches, represents an essential trade-off between time and accuracy: increasing the number of micro-batches slows the computation and increases the performance, and vice-versa.
Our configuration aimed to maximize the performance, which is already translated into a noticeable accuracy drop.
Target models with regularizers behave instead very similarly with respect to the baseline.

\begin{table}[t]
    \caption{The training execution times of the target model, expressed in seconds (lower is better).}
    \centering
    \begin{tabular}{|cccccccc|}
        \hline
        \textbf{  Dataset  } & \textbf{  Baseline  } & \textbf{  $\varepsilon$ = 2  } & \textbf{  $\varepsilon$ = 4  } & \textbf{  $\varepsilon$ = 8  } & \textbf{  L2  } & \textbf{  Dropout  } & \textbf{  L2+Dropout  }\\
        \hline
        CIFAR-10 & 73.3 & 3029.8 & 2923.1 & 2897.1 & 69.1 & $\mathbf{67.4}$ & 70.4\\
        MNIST & 39.5 & 1886.2 & 1870.2 & 1859.8 & 31.2 & $\mathbf{29.5}$ & 29.6\\
        F-MNIST & $\mathbf{77.4}$ & 3876.7 & 3952.1 & 3967.1 & 78.2 & 83.0 & 78.5\\
        \hline
    \end{tabular}
    \label{tab:dp_times}
\end{table}

\begin{table}[t]
    \caption{The AUC scores of the membership inference attack against the proposed models over the test set (lower is better).}
    \centering
    \begin{tabular}{|cccccccc|}
        \hline
        \textbf{  Dataset  } & \textbf{  Baseline  } & \textbf{  $\varepsilon$ = 2  } & \textbf{  $\varepsilon$ = 4  } & \textbf{  $\varepsilon$ = 8  } & \textbf{  L2  } & \textbf{ Dropout } & \textbf{  L2+Dropout  }\\
        \hline
        CIFAR-10 & 0.689 & $\mathbf{0.527}$ & 0.545 & 0.536 & 0.619 & 0.571 & 0.551\\
        MNIST & 0.738 & 0.585 & 0.598 & 0.583 & 0.567 & 0.614 & $\mathbf{0.555}$\\
        F-MNIST & 0.642 & $\mathbf{0.541}$ & 0.564 & 0.549 & 0.564 & 0.552 & 0.575\\
        \hline
    \end{tabular}
    \label{tab:auc}
    \vspace{-5mm}
\end{table}

Table~\ref{tab:auc} show the results of the membership inference attack against the proposed target models, expressed via the AUC metric.
As a first observation, all the models trained via DP-SGD achieved a score closer to the random guessing attack, meaning privacy preservation is effectively guaranteed. 
In detail, for $\varepsilon=2$ there is an average AUC reduction for the attacker of 13.86\%.
The most interesting aspect concerns the results from the target models with regularization.
In fact, all of them achieved better protection from the membership inference attack with respect to the baseline.
L2 works better where the DP-SGD is less effective, reaching even better performance than all the $\varepsilon$ over the MNIST dataset, and an average AUC reduction of 10.63\%.
Dropout also demonstrates competitive privacy guarantees, almost like DP-SGD ones, achieving an AUC average reduction of 11.07\%.
Combining the two regularization techniques improves the average score, reaching an AUC reduction of 12.93\%.
From these results, we argue that regularization techniques empirically demonstrate a protection level comparable with the one obtained via differential privacy optimizers.

\begin{figure}[t]
    \centering
    \includegraphics[width=1\textwidth]{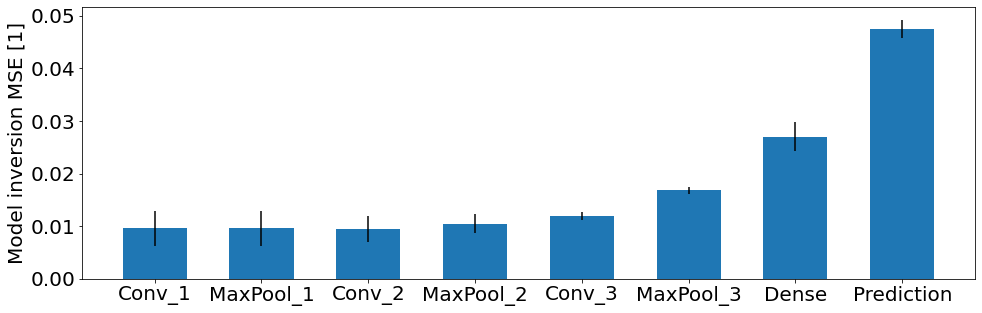}
    \caption{The average reconstruction MSE against the baseline target model over the test set. The scores are averaged over all the datasets (lower is better).}
    \label{fig:mse_baseline}
\end{figure}

\begin{table}[t]
    \caption{The variation of model inversion reconstruction MSE starting from each layer of the target model. The scores are expressed in percentages and are referred to the baseline model's score. The scores are averaged over all the datasets' test sets (higher is better).}
    \centering
    \begin{tabular}{|ccccccc|}
        \hline
        \textbf{  Layer  } & \textbf{  $\varepsilon$ = 2  } & \textbf{  $\varepsilon$ = 4  } & \textbf{  $\varepsilon$ = 8  } & \textbf{  L2  } & \textbf{  Dropout  } & \textbf{  L2+Dropout  }\\
        \hline
        Conv1       & 2.1   & 2.5   & -1.4  & -0.74              & $\mathbf{22.0}$   & 10.8\\
        MaxPool1    & 3.1   & 0.6   & 1.7   & -0.52              & $\mathbf{19.0}$   & -0.2\\
        Conv2       & 4.6   & 0.7   & 3.2   & -2.99              & $\mathbf{18.2}$   & 1.6\\
        MaxPool2    & 3.4   & -0.5  & -1.3  & -3.28              & $\mathbf{28.2}$   & 4.2\\
        Conv3       & 4.2   & -9.3  & -10.5 & -7.13              & $\mathbf{11.4}$   & 2.7\\
        MaxPool3    & -14.6 & -15.6 & -17.2 & -11.5              & $\mathbf{8.6}$    & -3.6\\
        Dense       & -21.7 & -25.8 & -25.2 & 17.1               & 17.3              & $\mathbf{21.3}$\\
        Prediction  & -17.7 & -18.2 & -15.8 & $\mathbf{73.8}$    & -1.8              & 64.2\\
        \hline
    \end{tabular}
    \label{tab:dp_layer_variation}
    \vspace{-5mm}
\end{table}

\begin{figure}[t]
    \centering
    {\includegraphics[width=.9\textwidth]{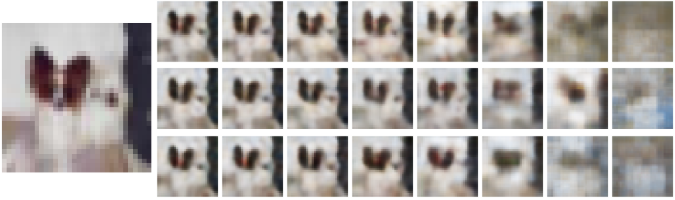}}
    {\includegraphics[width=.9\textwidth]{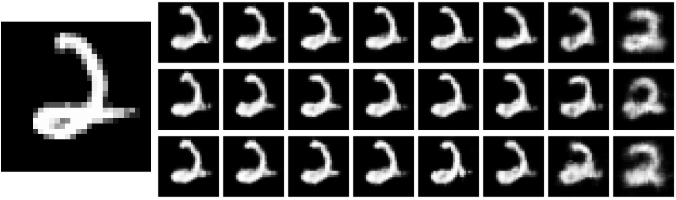}}
    {\includegraphics[width=.9\textwidth]{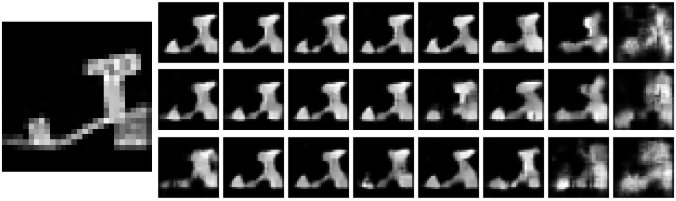}}
    \caption{Three examples of reconstructions from CIFAR-10 (top), MNIST (middle) and Fashion-MNIST (down). For each sample, the bigger image represents the ground truth to reconstruct. Each column j represents the model inversion reconstruction starting from the layer j of the target model (from left to right, Convolution, MaxPooling, Convolution, MaxPooling, Convolution, MaxPooling Flatten+Dense, Prediction). Each row corresponds to a different target model (from top to down, the baseline model, the ($\varepsilon=2$)-DP model and the l2+dropout model).}
    \label{fig:reconstruction_images}
\end{figure}

\begin{figure}[t]
    \centering
    \includegraphics[width=.47\textwidth]{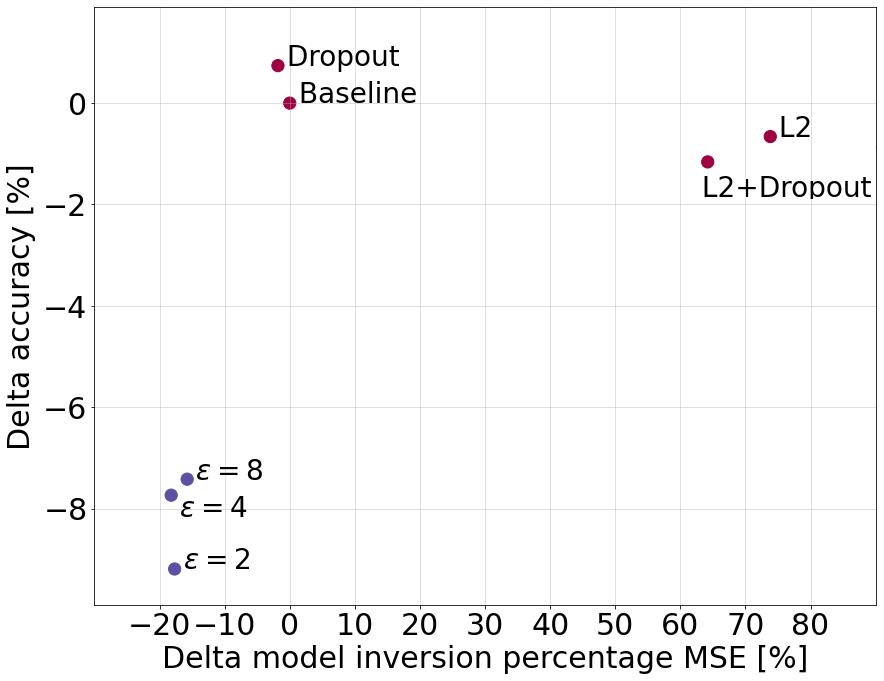}
    \includegraphics[width=.52\textwidth]{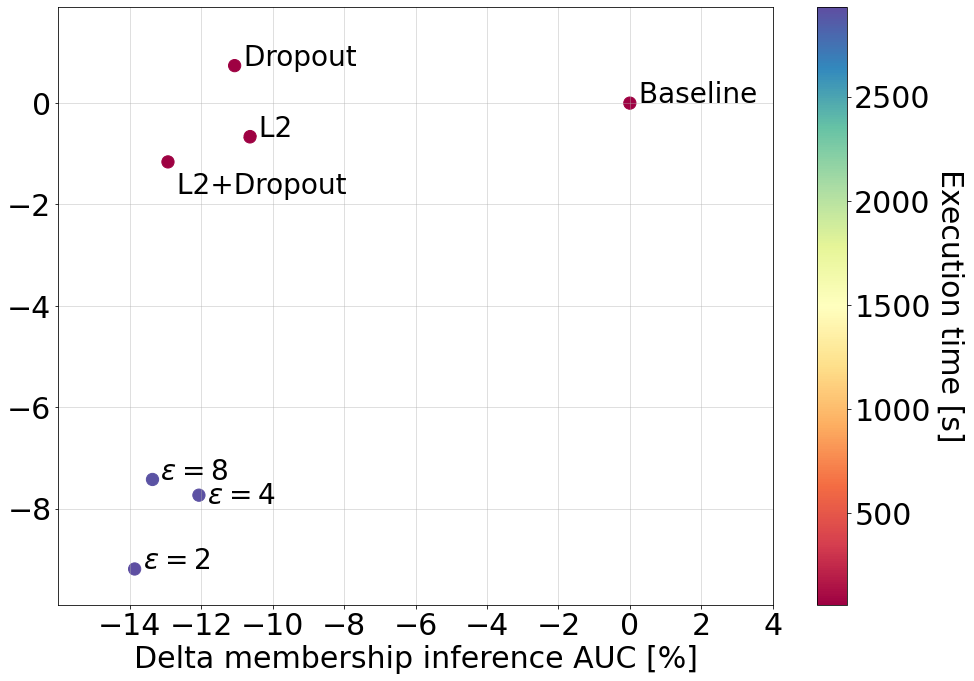}
    \caption{The final comparison showing the performance of all the proposed models in a black-box scenario, averaged over all the datasets. On the left, the x-axis represents the percentage MSE variation for model inversion attacks with respect to the Baseline model (higher is better), the y-axis represents the classification accuracy (higher is better). On the right, the x-axis represents the percentage AUC variation for membership inference attacks with respect to the Baseline model (lower is better), the y-axis represents again the classification accuracy (higher is better).}
    \label{fig:final_comparison}
\end{figure}

Figure~\ref{fig:mse_baseline} represents the results of the model inversion attacks against the baseline target model. 
Each bar corresponds to a layer and represents the reconstruction MSE averaged over all the datasets. 
This result shows that a model inversion attack is more effective as it is made closer to the network's input. 
It is reasonable given the growing amount of transformations that the attacker model would have to invert attacking from deeper layers.
It is interesting to note how the best case for an attacker in a white-box setting would be around four times more effective than the same attack done in a black-box setting, i.e., from the prediction layer.

Table~\ref{tab:dp_layer_variation} shows the percentage variations for each layer and each approach with respect to the baseline.
From these results, we notice an MSE reduction for all the final layers of the target models trained via DP-SGD.
A reconstruction error reduction means an advantage for the attacker, which implies, at least in this scenario, that differential privacy tends to improve the model inversion attack's quality rather than degrade it. 
Target models with regularization demonstrate completely different behavior.
We notice that l2 reduces the effects of model inversion in the layer on which it is applied, with an average improvement of the reconstruction error of 73.8\%.
Dropout has a similar effect on the target model, improving the protection of the layers before its application.
The main drawback is that it is impossible to apply dropout after the prediction layer.
By putting together l2 and dropout techniques, the obtained result is an average between the two, slightly improving the protection for hidden layers and increasing the defence in the prediction layer.

Figure~\ref{fig:reconstruction_images} shows three examples of a complete model inversion attack for the most significant target model configurations.
The proposed images show how the attacker network degrades its reconstructions as it starts to reconstruct the image from deeper layers.
In agreement with the numerical results, reconstructions from DP-SGD target model tend to present a lower artifacts amount with respect to the ones from the baseline or the combination of l2 and dropout regularization.

Finally, Figure~\ref{fig:final_comparison} summarizes all the results obtained so far.
The left image shows the percentage variations of the black-box model inversion MSE function of the accuracy variations. 
In the lower-left part are the worst results, in the upper-right the best.
Instead, the right image shows the percentage variations of membership inference AUC function of the accuracy variations.
In this case, the best region is the upper-left, while the worst is the lower-right.
The scatter colors represent the clear difference between the training time with and without DP-SGD optimizer.

Despite the undeniable benefits against membership inference attacks and the privacy measurability property, we can conclude that differential privacy techniques are not always a suitable option. In fact, in our scenario, this mechanism demonstrated poor quality trade-offs between protection guarantees, utility and training time.
Instead, the analyzed regularization techniques demonstrated promising performance. They are candidates for possible replacements in conditions where the training times and the required performances are subject to severe constraints.

\section{Conclusion}
This paper analyzed the benefits, drawbacks, and limitations of differentially-private optimization. 
In particular, we empirically showed how a differentially-private stochastic gradient descent optimizer could not always be considered a general protection paradigm for deep learning models despite its privacy budget guarantees.
Despite its effectiveness in protecting them against membership inference attacks, the mechanism degrades models' utility, requires much longer training times and increases model inversion attacks' quality.
We demonstrated how the combination of l2 and dropout regularization techniques is a valid protection alternative that does not degrade utility, requires contained training times, and provides a simultaneous resilience against membership inference and model inversion attacks.
\section*{Acknowledgment}
The European Commission has partially funded this work under the H2020 grant N. 101016577 AI-SPRINT: AI in Secure Privacy-pReserving computINg conTinuum.

\bibliographystyle{splncs04}
\bibliography{biblio}
\end{document}